\documentclass[manuscript,screen]{acmart}

\AtBeginDocument{%
  \providecommand\BibTeX{{%
    \normalfont B\kern-0.5em{\scshape i\kern-0.25em b}\kern-0.8em\TeX}}}

\usepackage[commandnameprefix=ifneeded]{changes}
\usepackage{xcolor}
\usepackage{color}
\usepackage{amsthm}
 \usepackage{mathrsfs}
 \usepackage{amsmath}
\usepackage{soul}

\newtheorem{myDef}{Definition}
\begin{document}

\title{Phased Deep Spatio-temporal Learning for Highway Traffic Volume Prediction}



\author{Weilong Ding}
\email{dingweilong@ncut.edu.cn}
\orcid{0000-0002-9982-5488}
\affiliation{%
\institution{School of Information Science and  Technology, North China University of Technology}
  \city{Beijing}
  \country{China}
  \postcode{100144}
  \streetaddress{No. 5 Jinyuanzhuang Road, Shijingshan District}
\institution{and Beijing Key Laboratory on Integration and Analysis of Large-scale Stream Data}
    \city{Beijing}
  \country{China}
  \postcode{100144}
  \streetaddress{No. 5 Jinyuanzhuang Road, Shijingshan District}
  }

\author{Tianpu Zhang}
\orcid{0000-0002-7146-8829}
\affiliation{%
\institution{School of Information Science and  Technology, North China University of Technology}
  \city{Beijing}
  \country{China}
  \postcode{100144}
  \streetaddress{No. 5 Jinyuanzhuang Road, Shijingshan District}
\institution{and Beijing Key Laboratory on Integration and Analysis of Large-scale Stream Data}
    \city{Beijing}
  \country{China}
  \postcode{100144}
  \streetaddress{No. 5 Jinyuanzhuang Road, Shijingshan District}
  }

\author{Zhe Wang}
\orcid{0000-0002-8223-1114}
\affiliation{%
  \institution{Cloud \& Smart Industries Group, Tecent Inc}
  \city{Beijing}
  \country{China}
 }


\begin{abstract}
 Inter-city highway transportation is significant for citizens’ modern urban life and generates heterogeneous sensory data with spatio-temporal characteristics. As a routine analysis in transportation domain, daily traffic volume estimation faces challenges for highway toll stations including lacking of exploration of correlative spatio-temporal features from a long-term perspective and effective means to deal with data imbalance which always deteriorates the predictive performance. In this paper, a deep spatio-temporal learning method is proposed to predict daily traffic volume in three phases. In feature pre-processing phase, data is normalized elaborately according to latent long-tail distribution. In spatio-temporal learning phase, a hybrid model is employed combining fully convolution network (FCN) and long short-term memory (LSTM), which considers time, space, meteorology, and calendar from heterogeneous data. In decision phase, traffic volumes on a coming day at network-wide toll stations would be achieved effectively, which is especially calibrated for vital few highway stations. Using real-world data from one Chinese provincial highway, extensive experiments show our method has distinct improvement for predictive accuracy than various traditional models,reaching 5.269 and 0.997 in MPAE and R-squre metrics, respectively.
\end{abstract}



\maketitle

\section{Introduction}
With the flourishing development of inter-city transportation, highway plays an important role for citizens in their urban modern life, and traffic congestion has become one of the most serious issues worldwide. Since highway is the enclosed environment, once congestion occurs, it will affect traffic seriously~\cite{article_1}. In fact, the capacity of road network has not been explored enough, and network-wide traffic control is imperative for official transportation guidance and personal travel planning. Being aware of the traffic is the first step to solve transportation problems~\cite{article_2}. As one of transportation domain fundamental measurements, traffic volume, i.e. the number of vehicles at given locations, reflects the highway traffic states. Accordingly, daily traffic volume prediction is for coming days has been adopted as a domain routine analysis to alleviate traffic congestion. Heterogeneous sensory data from distinct sources can be employed, and toll data at highway stations is often studied for that analysis. Here, the toll data keeps timestamps and locations when a vehicle was entering or exiting a station, which has exact locality with high quality~\cite{article_3}. 

Over last decade, many solutions have been studied extensively in the perspectives of statistics, machine learning, and deep neural network. However, it still faces challenges to predict daily traffic volumes in practice due to following inherent limitations. One the one hand, correlative spatio-temporal factors have not been fully considered in a relatively long term. Temporal feature is the only emphasis in early statistical solutions, and spatial proximity has been widely adopted in current methods. For example, downstream traffic instead of upstream in highway network influences more for future traffic volume at a location~\cite{article_4}. Implicit calendric periodicity (e.g., holidays or weekends) and external meteorological conditions (e.g., heavy snow or heavy rain) may affect traffic on certain days and bring similar traffic patterns~\cite{article_2} in highway. All those have been seldom considered comprehensively in current works yet. On the other hand, massive data with imbalanced distribution at toll stations deteriorates the performance of traffic prediction. Common machine learning models require normalization to avoid over-fitting~\cite{article_5}, but usually regard some great traffic volumes at “vital few” stations as outliers. It brings large errors to predict for those pivot locations in highway network. 

In this paper, a phased deep spatio-temporal learning method is proposed to predict daily traffic volume. It includes three phases: in feature pre-processing phase, data is normalized elaborately at toll stations in highway network; in spatio-temporal learning phase, a hybrid model combining fully convolution network (FCN) and long short-term memory (LSTM) is designed to capture temporal and spatial features; in decision phase, traffic volumes on a coming day at network-wide toll stations would be predicted effectively. Our contributions can be concluded as two aspects. (1) Comprehensively considering time, space, meteorological condition, and calendric factors from heterogeneous data, daily traffic volume prediction is effective in accuracy. (2) A three-phased hybrid model is adopted, combining convolution network and LSTM, can learn temporal and spatial characteristics more exactly. (3) Evaluated on the real-world data in a practical project, performance improvement and convincing benefits are proved by extensive experiments. Although discussed in a specific highway domain, our work is general to be employed in other domains for daily trends (e.g., city-wide passenger demand, crowd flow) prediction. 

The rest of this paper is organized as follows. Section 2 discusses related works; Section 3 shows preliminaries including motivation and problem definition; Section 4 elaborates our phased method for daily traffic volume prediction; Section 5 demonstrates performance and effects by experiments; Section 6 summarizes conclusion.

\section{Related work}
Highway traffic volumes are important in business, but their prediction faces challenges in performance and effects. We analyse recent studies, and divide those methods into three classes, each of which has its own merits. 

The first class belongs to statistical methods. Also known as parametric approaches~\cite{article_6}, those works focus on exploiting combinatorial optimization dependencies among multivariable factors to improve predictive precision~\cite{article_7}. Assuming stationary temporal or historical factors respectively, ARIMA (Autoregressive Integrated Moving Average model)~{\cite{article_8}} and HA (Historical Average)~{\cite{article_9}} are such common ways for traffic volume prediction. However, traffic data is always too complex to satisfy the assumptions above, so those models usually perform poorly on huge data in practice. Extended Kalman-filtering model~\cite{article_11} and ARIMA+~\cite{article_12} are employed for macroscopic and single-location traffic prediction. Due to the computational complexity of tuning parametric weights at limited locations, they are infeasible for prediction at network-wide stations. 

The second class lies in machine learning methods. Also known as non-parametric approaches, those non-linear models can flexibly present multiple features for traffic volume. For example, Wavelet model~\cite{article_13} and time delay neural network through genetic algorithm~\cite{article_14} are used for traffic volume prediction. But they focus on short-term trends within 30 minutes and cannot be applied directly for long term like one day due to different feature granularity. Other models such as KNN (K-Nearest Neighbor)~\cite{article_3}, SVR (Support Vector Regression)~\cite{article_15, article_16}, and ensemble gradient boosting~\cite{article_17, article_18} are also widely used after complex feature engineering. Most of them are only work well for major arterials locations, and perform poorly on other ones. Moreover, they are sensitive to the amount and quality of training data, and require specific data calibration before prediction.

The third class is deep learning (DL) methods. As a branch of machine learning, hybrid DL is popular nowadays due to their pretty high accuracy. To predict traffic trends, DL methods usually combine convolutional neural networks (CNN) with recurrent neural networks (RNN), in which the former is utilized to learn spatial dependency and the latter is for temporal dynamics~\cite{article_19, article_20}. In Euclidean space, traditional CNN (i.e., 2D CNN) captures two dimensions of features through convolutional operations on local neighborhoods with a 2D convolutional kernel. It has successfully applied in cities divided into regular grids by two dimensions of latitude and longitude~\cite{article_6, article_21, article_22, article_23}. However, it is hard for 2D CNN to learn spatio-temporal dynamics when more dimensions have to be considered. Accordingly, based on 2D CNN, efforts in four types have been made~\cite{article_23}, in which CNN combining RNN (or their variants) is the most widely adopted one. That is, to exact spatial feature, flattened dimensions (1D or 2D) of space are imported to CNN; to capture temporal dependencies of the output of CNN part, Long Short Term Memory (LSTM) and Gated Recurrent Unit (GRU), simplified variants of RNN, are commonly utilized. Such works like~\cite{article_24, article_25} have achieved well experimental results. They build temporal correlations on high-level spatial features, but lack low-level ones~\cite{article_23}. Besides, hybrid DL methods have been studied for urban trend prediction. For example, a merged LSTM model is proposed~\cite{article_26} on three different data sets to complete comparative analysis; a load forecasting approach is proposed~\cite{article_27}  for power systems through an ensemble of LSTM with full connected network. Moreover, Graph Convolutional Network (GCN), like~\cite{article_1, article_19}, is extensively adopted recently. For example, STFGNN~\cite{STFGNN} calculates the similarity of traffic sequences by DTW algorithm and stitches multiple graphs into a fusion graph based on similarity to obtain spatio-temporal dependence of the traffic sequences. ASTGCN~\cite{ASTGCN} extracts spatio-temporal features at different resolutions of traffic sequences independently and combines these saptio-temporal features in the final output layer to obtain the final traffic prediction results. Their popularity comes from better performance considering connectivity and globality~\cite{article_20}, but the overhead for complicated graph convolutions operation is not negligible either. 

In fact, it is hard to prove that one method is clearly superior over others in any situation. Although popular and mainstream in the literature of these years, various DL methods still remain open questions. DL is sensitive to the quality of training data and requires data pre-processing to avoid over-fitting. Data calibration or interpolation is necessary~\cite{article_28}, and data normalization is significant to standardize input data for specific model. Our work as a DL solution, data normalization is designed elaborately in feature pre-processing phase according to long-tail distribution of traffic volumes; moreover, daily traffic volume especially for vital few stations are calibrated logically through explainable statistical norms.

\begin{table}[]
\caption{The structure of a toll data record.}
\begin{tabular}{llll}
\hline
\hline
Notation         & Description                   & Example              & Type   \\
\hline
collector\_id    & toll collector identity       & XXXX080169           &        \\
vehicle\_license & vehicle identity              & XXDFH5XX             &        \\
vehicle\_type    & vehicle type                  & 1                    & Entity \\
card\_id         & vehicle passing card identity & 4101152822010XXXXXXX &        \\
etc\_id          & vehicle ETC card identity     & XXX7887              &        \\
etc\_cpu\_id     & ETC card chip identity        & XXX011               &        \\
\hline
entry\_time      & vehicle entry timestamp       & 2018/1/23 15:55:44   & Time   \\
exit\_time       & vehicle exit timestamp        & 2018/1/23 16:02:50   &        \\
\hline
entry\_station   & identity of entry station     & 33011                &        \\
entry\_lane      & lane number of entry station  & 3                    & Space  \\
exit\_station    & identity of exit station      & 33012                &        \\
exit\_lane       & lane number of exit station   & 1                    &       \\
\hline
\hline
\end{tabular}
\label{tab1}
\end{table}

\section{Preliminary}
\subsection{Motivation}
Our work originates from Highway Big Data Analysis System running in Henan, the most populated province in China. The system we built has been in production since October 2017 and is expected to improve routine business analytics for highway management through Big Data technologies. Operated by Henan Transport Department, a billion records of heterogeneous data in recent two years have been imported into the system, such as meteorological data, solar and lunar calendric data, real-time license plate recognition data, and toll data. A record of toll data is generated from highway toll station when a vehicle is passing. As the typical spatio-temporal data structure in spatial points during given time slots. Due to time duration, traffic volume prediction can be classified into short-term, medium-term and long-term~\cite{article_32, article_33} prediction. Our previous works~\cite{article_3, article_18} have been applied in the system for short-term traffic volume prediction and some analytics like potential hot-spot detection. Daily traffic volume prediction, one of typical long-term prediction, is completely studied in-depth in this paper. 

As Table~\ref{tab1}, a record contains 12 attributes including six entity attributes, two temporal attributes and four spatial attributes. Traditionally, toll data from sensors would be loaded into a data warehouse at the end of a day. After ETL (Extract, Transform, Load) step with necessary pre-processing like~\cite{article_29}, multiple business analytics would run on massive and heterogeneous data. As one significant analytics, traffic volume prediction is to estimate future traffic volumes. Here, traffic volume, also termed as traffic flow in other workarounds like~\cite{article_1, article_6, article_24, article_28, article_30, article_31}, counts vehicles passing specific normalization is designed elaborately in feature pre-processing phase according to long-tail distribution of traffic volumes; in deep learning phase, input is mapped into proper Euclidean space to feed hybrid model combining CNN and LSTM for better predictive accuracy.

\subsection{Problem definition}
Daily traffic volume of toll stations in whole highway network can be formally defined as follows, according to bi-direction traffic at a given toll stations.
\begin{myDef}
    \textbf{Daily exit (entry) traffic volume.} Daily exit (entry) traffic volume presented as $xTF_l^d (nTF_l^d)$ counts the vehicles $V$ exiting (entering) toll station $l\in{\mathcal{L}}$ in a day $d$. Hence, network-wide daily exit traffic volume can be described as  $xTF_l^d=[xTF_1^d,xTF_2^d,...,xTF_l^d,...,xTF_L^d]$. Here, $L$ is the cardinality of toll stations $\mathcal{L}$ in highway. In this paper, traffic volume is the exit one by default if no other emphasis. The exit traffic volume rather than the entry one is main focus in domain because tolls would be charged only when vehicles exiting a station. 

    Accordingly, the problem focused in this paper can be abstracted as the following definition.

\end{myDef}
\begin{myDef}
    \textbf{Network-wide prediction of daily traffic volume.} On a current day $d$, the input feature $X\in{\mathbb{R}^{L*H*N}}$  can be represented as Equation~\ref{equation_1}, where $H$ is the length of past continuous days $\mathcal{H}$. A row $X_l\in{\mathbb{R}^{N*N}}$ is the feature of a toll station $l$ on $H$ days; a column $X^h\in{\mathbb{R}^{L*N}}$ is the feature on a day $h\in{\mathcal{H}}$ for $L$ toll stations. Given the input $X$ on a recent day $d$, the objective is to predict the traffic volume of any toll station on the coming day $d+1$. Hence, network-wide prediction of traffic volume on day $d+1$ can be described as Equation~\ref{equation_2} through a required model $F$. 
\begin{equation}
    X=\begin{bmatrix} X_1 \\ \cdots \\X_l \\ \cdots \\X_L \end{bmatrix}=\begin{bmatrix} X^{d-H+1} & \cdots & X^H & \cdots& X^{d-1} & X^d\end{bmatrix}=[X^h_l]^{L*H}
\label{equation_1}
\end{equation}

\begin{equation}
    xTF^{d+1}_{\mathcal{L}}=F(X)
\label{equation_2}
\end{equation}

    Here, given any day $h\in{\mathcal{H}}, X(l,h)=X_l^h\in{\mathbb{R}^N}$ is the N-dimensional feature of traffic volume at toll station $l$.
    
\end{myDef}

\section{Phased Deep Spatio-temporal Learning Method}

\subsection{Methodology and feature pre-processing}
\begin{figure}[h]
  \centering
  \includegraphics[width=0.5\textwidth]{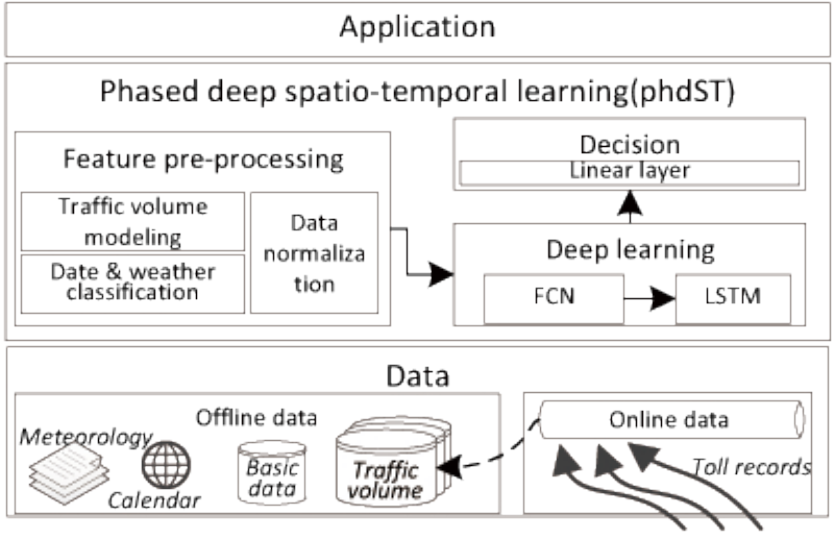}
  \caption{Framework of our method.}
  \label{fig_1}
\end{figure}

In highway domain, traffic volume prediction is widely used to find hotspots and estimate recent status in various temporal or spatial granularities. In this paper, we focus on daily traffic volume prediction according to Definition 1~2, and propose phdST (phased deep spatio-temporal learning) method. The framework of our method is presented as Figure~\ref{fig_1} with its input and output. Our method as a routine analysis would execute once a day at 12:00 a.m. on the data of recent three months, and output predicted results at network-wide toll stations on a coming day.

The input of phdST is online and offline data. Raw records of toll data are received continuously through a message broker, and then aggregated as traffic volume data into No-SQL database. Related procedures like data cleaning and aggregative calculation can be referred in our previous works~\cite{article_3,article_29}. External calendar and meteorology data are extracted periodically from dedicated data sources, and then stored into a relational database. Business basic data, such as profiles of station, section and highway line, has been imported in that relational database. On such heterogeneous data, phdST includes three phases. In feature pre-processing phase, external data (i.e., date and weather) is labelled by classification according to domain rules, and traffic volume would be normalized elaborately. In spatio-temporal learning phase, a hybrid model is introduced: spatial feature of heterogeneous data is fused by FCN, and then temporal feature is captured collaboratively by LSTM.  In decision phase, the traffic volume of the whole network would be divided into two groups according to the traffic size to eliminate the impact of skewness distribution on the prediction accuracy and then traffic volumes at network-wide toll stations would be calibrated for coming-day assisted with statistical norms. Such predicted results would be written back into the relational database. The output applications can employ those results to complete business requirements, such as analytical visualization and potential hotspots discovery.

In our phdST method, feature pre-processing to build feature is the first phase of the threes. Based on Definition 1~2 and the observations in Section 3.1, feature of traffic volume (i.e., exit traffic volume) can be defined as follows with dimensions of time (e.g., recent values at a station), space (e.g., values at adjacent stations), date (e.g., workday or other days), and weather (e.g., extreme condition or not). 
\begin{myDef}

    \textbf{Feature of exit traffic volume.} On a day $h$ at station $l$, feature of exit traffic volume is represented as $X^h_l=(W^h_l, D_h, nTF^h_{l'_1}, \cdots, nTF^h_{l'_j}, \cdots, nTF^h_{l'_\eta}), j=1\cdots \eta-1, \eta\in{\mathbb{Z^+}}$. Three dimensions exist in $X^h_l$: the first $W^h_l\in{0,1}$  is weather category at location $l$ on day $h$; the second part $D_h\in{0,1,2}$ is date category of day $h$; the third is entry traffic volumes of $\eta$ upstream dependent stations~\cite{article_3} $l'_1, l'_2, \cdots, l'_\eta$ on day $h$. That is, the dimensionality $N$ of Definition 2 would be $N=\eta+2$. These factors are discussed below.
\begin{equation}
   W_l^h=\left\{
   \begin{array}{lr}
        1, if ((visibility|rain|snow).level\geq2)  V ((wind|temperature).level\geq 3),&\\
        0, otherwise.&\\
    \end{array}
    \right.
\label{equation_3}
\end{equation}
    (1) In the meteorological dimension, extreme weather condition (i.e., $W_l^h =1$) is defined by Equation~\ref{equation_3} under domain standard~\cite{article_34}. Meteorology data of counties and cities has been modelled as label-encoding $W_l^h$(visibility, rain, temperature (low, high), wind, snow) at $l\in{ \mathcal{L}}$ on $h\in{\mathcal{H}}.$
    
    (2) In the calendric dimension, holidays and weekends are distinguished from others according to solar and lunar calendars. Referred to~\cite{article_32}, $D_h$ is presented as Equation~\ref{equation_4}.
    
    \begin{equation}
       D_h=\left\{
       \begin{array}{lr}
            1, if h is a weekend,&\\
            2, if h is a holiday,& \\
            0, otherwise.&\\
        \end{array}
        \right.
    \label{equation_4}
    \end{equation}

    (3) In the spatial dimension $nTF^h_{l'_j}$, entry traffic volumes of $\eta$ upstream dependent stations are considered on day $h$, because a vehicle exiting a location must have entered highway in an upstream one. Such an upstream dependent location is referred to our previous work~\cite{article_3} and implies spatial topological correlation of highway network in a statistical view: it is found by ascending order after sorting absolute differences between cartographic distance $(l, l'_j)$ and vehicles’ average mileage at given $l$.
    
    Then, data normalization on traffic volumes of Definition 3 is required to re-scale values to a notionally common range. To emphasize vital few values of long-tail distribution rather than discarding them as outliers, we adopt Box-Cox transformation~\cite{article_35} in our normalization strategy here. To realize end-to-end prediction through DL model, the inverse transformation would also be used in decision phase. Moreover, a binary classification is employed to distinguish ``vital few`` stations from others in long-tail distribution. The toll stations, whose volume is more than three times than that standard deviation, are regarded as vital few ones according to three-sigma rule~\cite{article_36}. For any station $l$ of those ones, a dedicated flag $vital_l=1$ is labeled then.
    
    Finally, in the feature pre-processing phase, feature is built for further usage.
    
\end{myDef}

\subsection{Spatio-temporal learning through a hybrid model}

After feature pre-processing, feature would be feed as input of spatio-temporal learning phase. The neural network structure here is illustrated in Figure 2 including four layers.

\textbf{Feature splitting layer} is to divide input feature tensor by days. According to the notations in Definition 2, feature $X$ would be split into $H$ columns and each column $X^h\in\mathbb{R}^{L\times N}$ is a two-dimensional matrix. Here, $h$ is any day between $d-H+1$ and current day $d$.

\begin{figure}[h]
  \centering
  \includegraphics[width=0.85\textwidth]{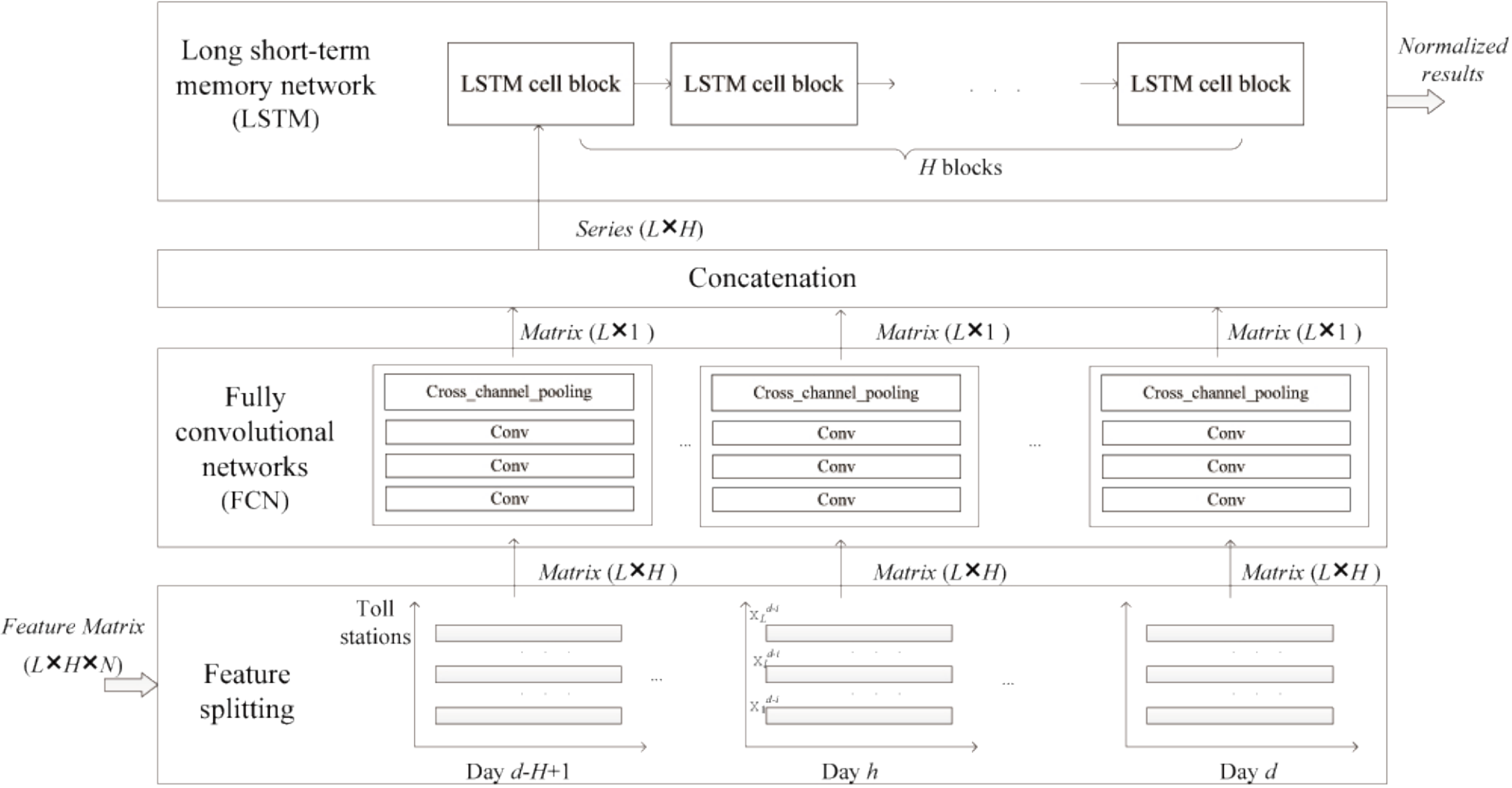}
  \caption{Network structure in spatio-temporal learning phase.}
  \label{fig_2}
\end{figure}

\begin{figure}[htbp]
\centering
\begin{minipage}[t]{0.48\textwidth}
\centering
\includegraphics[width=6cm]{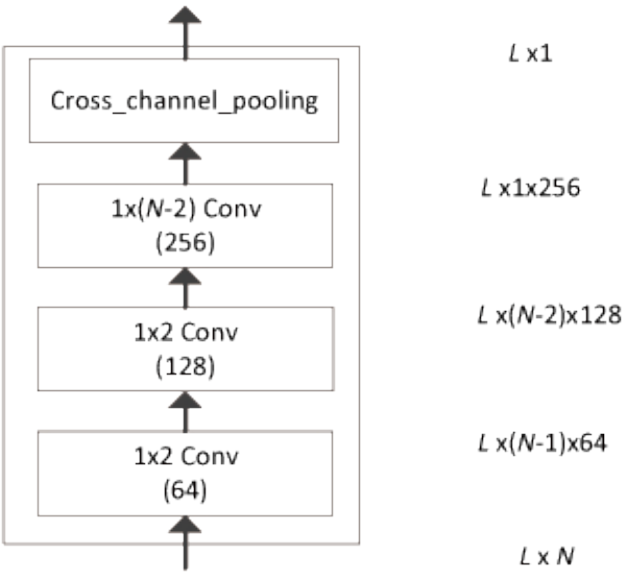}
\caption{FCN module structure.}
\label{fig_3}
\end{minipage}
\begin{minipage}[t]{0.48\textwidth}
\centering
\includegraphics[width=6cm]{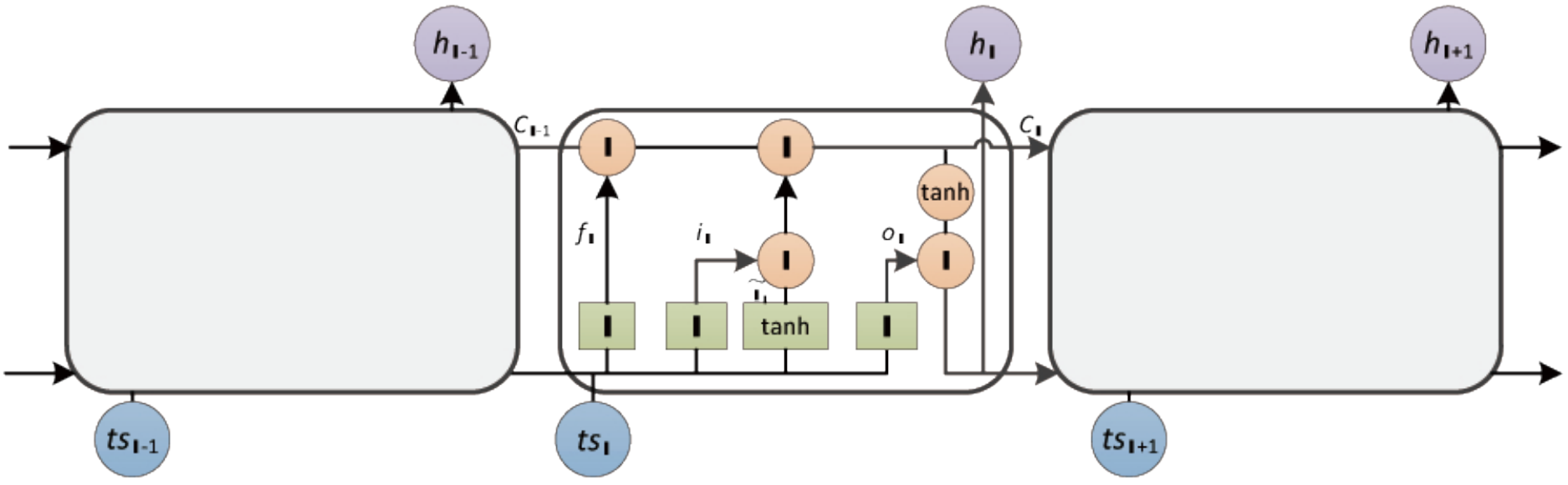}
\caption{LSTM block structure.}
\label{fig_4}
\end{minipage}
\end{figure}

\textbf{FCN layer} contains $H$ to modules, each of which is a fully convolutional network to process an output from feature splitting layer. Through up-sampling technology~\cite{article_37}, end-to-end pixel-wise prediction can increase the feature dimensions to keep output resolution as input after convolution. We borrow such idea in FCN modules with hetero convolution kernels, and compose three convolutional operations (Conv) with a pooling operation in the end. These operations are represented as Equation~\ref{equation_5} and \ref{equation_6} respectively. The detailed structure of FCN module is illustrated in Figure~\ref{fig_3}. Here, the input is  $X^h\in\mathbb{R}^{L\times N}$. As Definition 3, feature $X_l^h$ (i.e., a row of $X^h$ ) has $N$ dimensions where the first is meteorological category, the second is calendric category, and the others are entry traffic volumes. In a FCN module, after the first Conv with a $1\times2$ kernel in $64$ channels, up-sampling brings a $L\times(N-1)\times64$ result to fuse the first dimension into rest ones. Likely, the second Conv with a $1\times2$ kernel in $128$ channels fuses meteorological and calendric dimensions into rest ones, and brings a $L\times(N-2)\times128$ result. By such hetero kernels, the first two label-encoding dimensions would be melted into other dimensions of normalized traffic volumes. The third Conv with a $1\times(N-2)$ kernel in {$256$} channels merges the other dimensions, and leads a $L\times1\times256$ result. Here, ReLU is used as the activation function of these Convs. In order to reduce dimensions for those convolutional results, a cross channel pooling operation~\cite{article_38} is adopted to uniformly perform average pooling on $256$ channels, as Equation~\ref{equation_6}. Eventually, a $L\times1$ output is generated from a $L\times N$ input in a FCN module. 

\begin{equation}
   Conv_i=\left\{
   \begin{array}{lr}
        ReLU\left(Convolution\left(X^h\right),kernel\right),i=1&\\
        ReLU\left(Convolution\left({\rm Conv}_{i-1}\right),kernel\right),\ i=2,3& \\
    \end{array}
    \right.
\label{equation_5}
\end{equation}

\begin{equation}
    {\rm FCN}_h=cross\_channel\_average\_pooling({\rm Conv}_3)
\label{equation_6}
\end{equation}

\textbf{Concatenation layer} gathers $H$ outputs from FCN layer, concatenates them into a series of $H$ components ordered by date. Each of the components in series is a $L×1$ column vector. In LSTM layer, a neural network composed by $H$ LSTM blocks is to learn temporal characteristics, and each block would receive one component of series from concatenation layer. The structure of such a block is presented in Figure~\ref{fig_4} to remove or add information to cell state by multiple gates. To optionally let information through, gates are composed of a sigmoid operation $\sigma$ and a point-wise multiplication operation $×$, and three gates are included in a block. Forget gate $f_{d`}$ presented as Equation~\ref{equation_7} decides what information to throw away from cell state, considering current input component and previous output $h_{h-1}$. As Equation~\ref{equation_8}, input $i_{d`}$ gate decides what new information to store in cell state. After a new vector  $\widetilde{C_{h}}$ for addition is created, the new cell state $C_h$ is updated. As Equation~\ref{equation_9}, output gate $o_{d`}$ decides what to output. After creating $o_{d`}$, output $h_h$ is determined by multiplying it by a $tanh$ operation on cell state. Here, the notation * in following equations denotes an element-wise product. 

By serializing feature into data series over $H$-size features, LSTM layer employs normalized traffic volumes as labels for learning. The predicted network-wide traffic volumes as normalized results are generated eventually. 

\begin{equation}
f_{d`}=\sigma(W_f\left[h_{h-1},x_h\right]+b_f)
\label{equation_7}
\end{equation}

\begin{equation}
i_{d`}=\sigma\left(W_i\left[h_{h-1},x_h\right]+b_i\right);\ \widetilde{C_h}=\tanh{\left(W_C\left[h_{h-1},x_h\right]+b_C\right)};C_h=f_h\ast C_{h-1}+i_h\ast\widetilde{C_h}
\label{equation_8}
\end{equation}

\begin{equation}
o_{d`}=\sigma\left(W_o\left[h_{h-1},x_h\right]+b_o\right);h_h=o_{d`}\ast\tanh(C_h)
\label{equation_9}
\end{equation}

\subsection{Decision to calibrate predicted results}
\begin{figure}[h]
  \centering
  \includegraphics[width=0.75\linewidth]{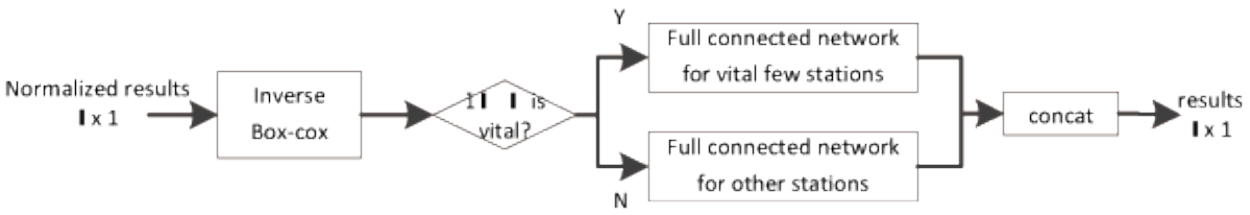}
  \caption{Data flow in decision phase.}
  \label{fig_5}
\end{figure}

After deep learning phase, results at network-wide toll stations are achieved as normalized values. If directly transformed by the inverse Box-Cox, the predictive effects of vital few stations in long-tail distribution cannot fit ground true well as expected because their absolute values are too large. Therefore, in the last decision phase, a linear layer is designed to decide how to calibrate those results by inverse normalization. The data flow of linear layer in decision phase is illustrated as Figure~\ref{fig_5}.

The basic idea here is to classify results from previous phase into two groups and a respective model on each group is used. One group is the normalized results of vital few stations, and the other is that of other stations. Before the classification, inverse Box-cox operation is calculated on those normalized results. It is used to transform the range of those values back to that of original traffic volumes. Then, the network-wide stations $\mathcal{L}$ are classified according to the dedicated flag mentioned in Section 4.1: for any $l\in{\mathcal{L}}$, its normalized results would be put into a corresponding group by flag $vital_l$ (1 or 0). That is, the vital few stations would be considered independently apart from others.

For either group, a fully connected neural network is trained to achieve the end-to-end predictive values of traffic volumes. The results of inverse operation fit to the ground truth with a linear relation. Using real traffic volumes as labels and the output of inverse Box-Cox operation as input, a fully connected network is composed by two hidden layers with an activation function ReLU in each layer. Therefore, two neural networks are employed respectively on each data group in the same way. After the concatenation from the outputs of two networks, the final predicted traffic volumes at network-wide toll stations would be achieved then. 

\section{Evaluation}
\subsection{Setting}
In the project mentioned in Section 3.1, our method is evaluated by three experiments. An Figure~\ref{fig_1}, to maintain raw toll data and aggregative traffic volume data of input data layer, three virtual machines of our private Cloud form a HBase 1.6.0 cluster, each of which owns 4 cores CPU, 22 GB RAM and 700 GB storage. Toll data of Henan highway is generated as the speed of 1.5 million records per day from 274 toll stations (i.e., $L= 274$). Another virtual machine (4 cores CPU, 8 GB RAM and 200 GB storage installing CentOS 6.6 x86\_64 operating system) is used to install MySQL 5.6.17 as the relational database for both business profiles (station, section and highway line) and external data (i.e., calendric data and meteorological data). An Inspur rack server is used to deploy our phdST method, which owns 8 processors (Intel Xeon E5-4607 2.20GHz), 64 GB RAM, 80 TB storage, and two NVIDIA 1080 GPUs. 

Our method is implemented by Oracle JDK 1.7.0 and an open source deep learning framework TensorFlow 1.12.0. The training dataset of historical data is since June to August 2017 (i.e., $H=92$). The data in September 2017 is used as test dataset for prediction. Similar to our previous work~\cite{article_3}, we set dependent upstream parameters $\eta=3$ (i.e., $N=2+\eta=5$). 

To evaluate prediction effects, three metrics, mean absolute percentage error (MAPE), root mean square error (RMSE), and R-square, are employed respectively as Equation~\ref{equation_10} -~\ref{equation_12}. Here, on a given day at a toll station with sequential index $i\in\mathbb{N}$, ${\hat{y}}_i$ represents predicted value, $y_i$ represents real value and $n$ is the size of test dataset.

\begin{equation}
MAPE=\frac{100\%}{n}\sum_{i=1}^{n}\left|\frac{{\hat{y}}_i-y_i}{y_i}\right|
\label{equation_10}
\end{equation}

\begin{equation}
RMSE=\sqrt{\frac{1}{n}\sum_{i=1}^{n}(\hat{y}_i-y_i)^2}
\label{equation_11}
\end{equation}

\begin{equation}
R-square=1-\frac{\sum_{i=1}^{n}\left(y_i-{\hat{y}}_i\right)^2}{\sum_{i=1}^{n}\left(y_i-\bar{y}\right)^2}
\label{equation_12}
\end{equation}

\subsection{Experiment}
We design experiments below to compare the prediction results in two types of models.

\textbf{Experiment 1: Prediction effects comparison with statistical machine learning models.} In order to quantitatively evaluate predictive effects of our method, four statistical models, i.e., ARIMA, SVR, KNN and GBRT, are implemented for comparison by toolkit scikit-learn 0.20.3 in the same environment. ARIMA (Auto-Regressive Integrated Moving Average)~\cite{article_12} as a linear model is achieved with its optimal parameters $p=2$, $d=1$ and $q=3$. SVR (Support-Vector Regression)~\cite{article_39} as a non-linear model is used with its optimal parameters kernel=’rbf’, $C=500$, and $\epsilon=0.8$. KNN~\cite{article_3} as a classic non-parametric model is gained with the parameters setting $k=5$, $\theta = 10$ and $\eta =3$. Gradient boost regression tree (GBRT)~\cite{article_18} as an ensemble learning model is used with the best parameters setting $M=3000$, $d=3$, $r=0.5$, $N=120$, and $H=30$. All the three metrics above through those models are counted after predictions on the same test set. We conduct predictive comparisons through those models from two perspectives. One shows network-wide traffic volumes on six specific days; the other presents the traffic volumes of four typical stations on days of continuous two weeks. Note that, ARIMA and SVR have to be evaluated only in the latter perspective because a prediction only works at a single toll station. 

\begin{figure}[h]
  \centering
  \includegraphics[width=0.75\linewidth]{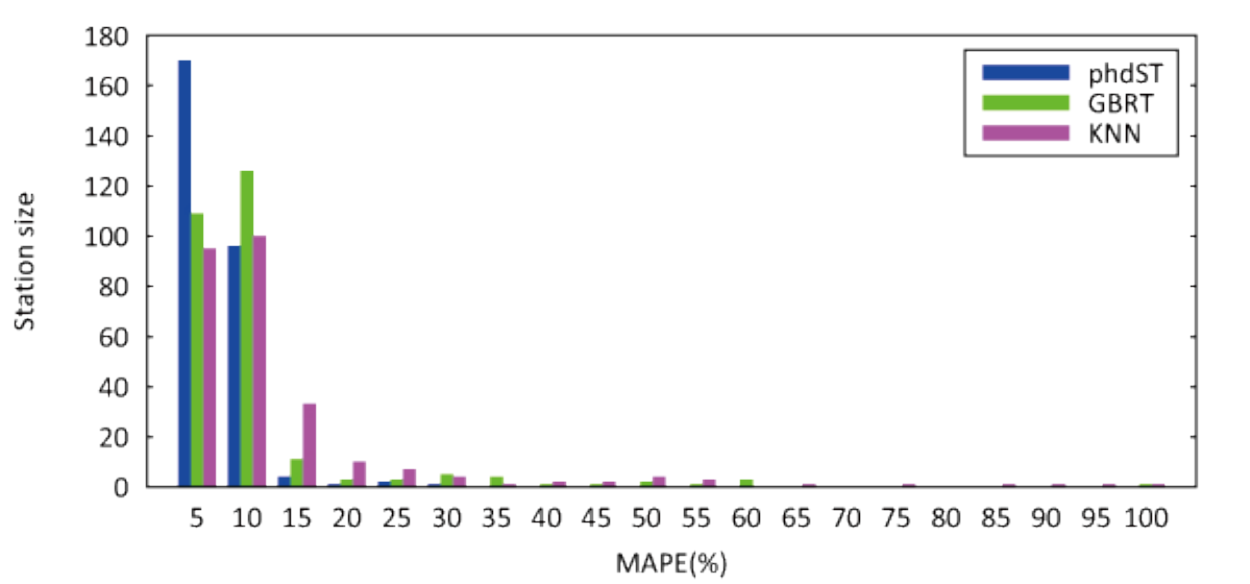}
  \caption{Data flow in decision phase.}
  \label{fig_6}
\end{figure}

The results from network-wide perspective are illustrated in Figure~\ref{fig_6} and Table~\ref{tab_2} for three models. At any toll station, the average of metric MAPE over the days in test data is calculated, and its distribution histogram is drawn as Figure~\ref{fig_6}. The toll station, whose average is not larger than the value of x-coordinate, would be counted as the accumulative value of y-coordinate. Two interesting evidences are found here. First, through any of the three models, the predictive results are with relatively low errors. Maxima of MAPEs at those toll stations are not larger than 15\%. Second, our method performs the best due to visible positive-skew distribution. Through our method, the histogram’s first bucket (i.e., with the lowest error) has largest count; the first two buckets (i.e., with MAPE smaller than 10\%) contains about 97\% of all the toll stations; no toll station has its MAPE larger than 30\%. GBRT performs a little better than KNN, but both have several toll stations whose MAPE is larger than 50\%. In fact, the vital few stations in long-tail distribution are such ones with bad accuracy. Furthermore, six specific days are used for detailed statistics as Table~\ref{tab_2}. The first two days are regular weekends, and the others are the last four days before a 7-day holiday Chinese National Day. On any of the six days, GBRT performs a little better than KNN on any day, and all the three models perform steadily in metrics MAPE and R-square with respective order of magnitude. Our phdST performs the best of all: metric MAPE is not large than 5.4, and the metric R-square is round 0.99. It comes from phased deep learning for traffic volume prediction. The combination of pre-processing which enables our method to ignore the side effect of data unbalance and application of multidimensional features including meteorology, calendar and spatio-temporal traffic enhance the accuracy of prediction results obtained by our method.

\begin{table}[]
\caption{Prediction performance from network-wide perspective.}
\begin{tabular}{c|ccc|ccc|ccc}
\hline
         & phdST    & GBRT    & KNN     & phdST    & GBRT    & KNN     & phdST    & GBRT    & KNN     \\
\hline
         & \multicolumn{3}{c|}{20170923} & \multicolumn{3}{c|}{20170924} & \multicolumn{3}{c}{20170927} \\
\cline{2-10}
MAPE(\%) & 5.269    & 10.246  & 15.155  & 5.394    & 12.6316 & 18.765  & 5.119    & 12.569  & 18.506  \\
RMSE     & 236.520  & 421.834 & 527.062 & 257.399  & 451.043 & 560.022 & 250.131  & 453.426 & 567.744 \\
R-square & 0.996    & 0.988   & 0.982   & 0.996    & 0.988   & 0.981   & 0.996    & 0.988   & 0.981   \\
\hline
         & \multicolumn{3}{c|}{20170928} & \multicolumn{3}{c|}{20170929} & \multicolumn{3}{c}{20170930} \\
\cline{2-10}
MAPE(\%) & 5.113    & 10.788  & 15.5906 & 4.529    & 9.972   & 14.279  & 5.248    & 10.968  & 15.980  \\
RMSE     & 219.126  & 515.936 & 642.547 & 211.379  & 577.472 & 728.854 & 239.917  & 406.793 & 503.736 \\
R-square & 0.997    & 0.986   & 0.978   & 0.998    & 0.984   & 0.974   & 0.996    & 0.988   & 0.981 \\
\hline
\end{tabular}
\label{tab_2}
\end{table}

\begin{figure}[h]
  \centering
  \includegraphics[width=\linewidth]{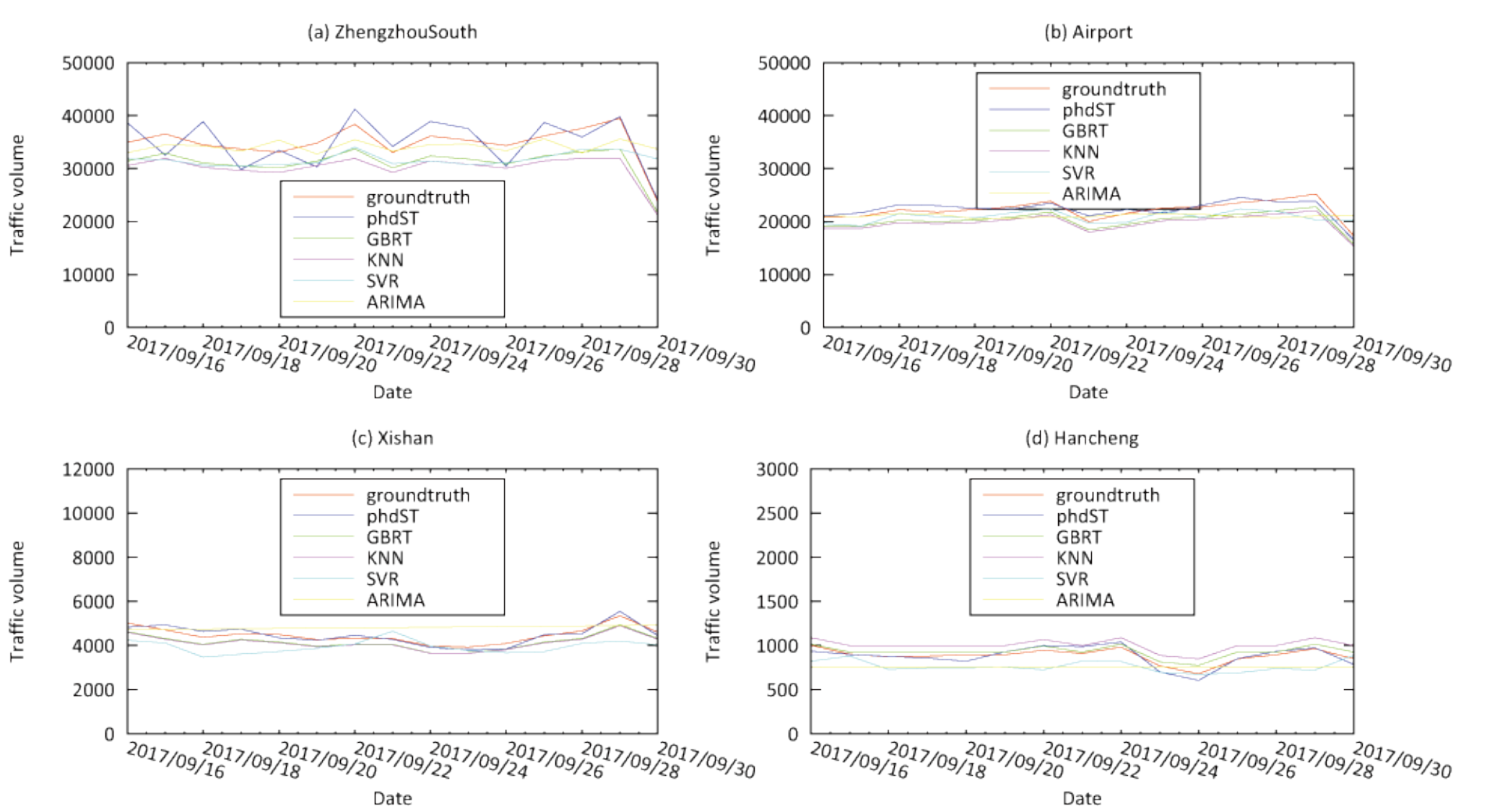}
  \caption{Traffic volume from station perspective.}
  \label{fig_7}
\end{figure}

The results from station perspective are demonstrated in Figure~\ref{fig_7} and Table~\ref{tab_3}. Four typical toll stations are chosen to illustrate their predictive effects on continuous 15 days. ZhengzhouSouth is the only station whose daily traffic volumes larger than 30 thousands; Airport is another top-5 station whose daily traffic volumes are around 20 thousands; Xishan represents majority stations whose traffic volumes are about five thousands; Hancheng is a small stations with daily traffic volumes fewer than one thousand. As shown in Figure~\ref{fig_7}, at each of those four toll stations, predicted traffic volumes through all the five models fit the result to ground truth. Compared those results with ground truth, phdST appears the most similar fluctuations of daily traffic volumes, and ARIMA cannot fit the fluctuation of the toll stations owning small traffic volumes. Further details can be found from the statistics in Table~\ref{tab_3}. According to three metrics, five models perform differently at stations. GBRT and SVR seem to perform well at stations of small traffic volumes, KNN achieves good results only at the stations of moderate traffic, and ARIMA cannot work well at toll stations with small traffic volumes. Our method performs the best at any toll stations with steady performance, which has MAPE smaller than 7.4 and R-square round 0.9. Extensive feasibility is proved due to such high predictive accuracy from phased deep learning.

\begin{table}[]
\caption{Prediction performance from station perspective.}
\begin{tabular}{c|ccccc|ccccc}
\hline
         & phdST    & GBRT      & KNN      & SVR      & ARIMA    & phdST   & GBRT     & KNN      & SVR      & ARIMA    \\
\hline
         & \multicolumn{5}{c|}{ZhengzhouSouth}                    & \multicolumn{5}{c}{Airport}                         \\
\cline{2-11}
MAPE(\%) & 7.297    & 10.344    & 13.171   & 12.310   & 7.315    & 3.364   & 8.732    & 10.920   & 7.484    & 7.860    \\
RMSE     & 2948.134 & 3731.4594 & 4773.283 & 4374.846 & 3343.994 & 813.132 & 1946.829 & 2435.598 & 1977.165 & 2223.104 \\
R-square & 0.952    & 0.924     & 0.875    & 0.895    & 0.939    & 0.987   & 0.926    & 0.884    & 0.924    & 0.904    \\
\hline
         & \multicolumn{5}{c|}{Xishan}                            & \multicolumn{5}{c}{Hancheng}                        \\
\cline{2-11}
MAPE(\%) & 3.369    & 6.927     & 7.585    & 12.345   & 10.322   & 4.997   & 5.354    & 13.579   & 13.996   & 15.243   \\
RMSE     & 168.671  & 314.651   & 344.589  & 642.133  & 504.297  & 51.191  & 50.756   & 119.789  & 143.830  & 150.818  \\
R-square & 0.985    & 0.949     & 0.939    & 0.789    & 0.870    & 0.972   & 0.972    & 0.845    & 0.777    & 0.754  \\
\hline
\end{tabular}
\label{tab_3}
\end{table}

\textbf{Experiment 2: Prediction effects comparison with deep learning models.} Besides phdST, three other deep learning models, i.e., LSTM, GCN, and GTM, are implemented for comparison in the same environment. LSTM as a variant of recurrent neural network has been widely used for time-series prediction. GCN is a graph convolutional model, which builds graph based on physical highway network~\cite{article_40}. GTM is a model composing GCN with LSTM to improve accuracy in temporal dimension~\cite{article_41}. All the three metrics above through those models are counted after predictions on the same test set. As the same setting of Experiment 1, we conduct predictive comparisons through those models from two perspectives. One shows network-wide traffic volumes on six specific days; the other presents the traffic volumes of four typical stations on days of continuous two weeks. Note that, LSTM has to be evaluated only in the latter perspective because a prediction only works at a single toll station.

\begin{figure}[h]
  \centering
  \includegraphics[width=0.75\linewidth]{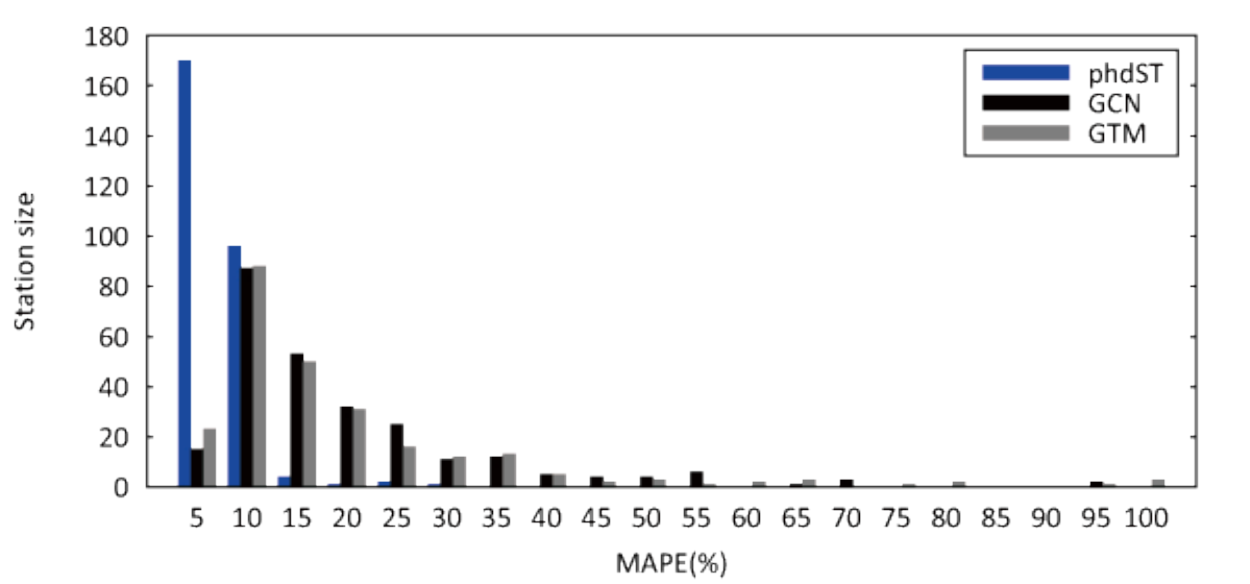}
  \caption{MAPE distribution from network-wide perspective.}
  \label{fig_8}
\end{figure}

\begin{table}[]
\caption{Prediction performance from network-wide perspective.}
\begin{tabular}{c|ccc|ccc|ccc}
\hline
         & phdST    & GCN     & GTM     & phdST    & GCN     & GTM     & phdST   & GCN      & GTM      \\
\hline
         & \multicolumn{3}{c|}{20170923} & \multicolumn{3}{c|}{20170924} & \multicolumn{3}{c}{20170927}  \\
\cline{2-10}
MAPE(\%) & 5.269    & 25.125  & 28.719  & 5.394    & 26.874  & 34.673  & 5.119   & 19.627   & 28.106   \\
RMSE     & 236.520  & 737.704 & 504.541 & 257.399  & 610.625 & 529.441 & 250.131 & 582.001  & 529.192  \\
R-square & 0.996    & 0.964   & 0.983   & 0.996    & 0.978   & 0.983   & 0.996   & 0.980    & 0.984    \\
\hline
         & \multicolumn{3}{c|}{20170928} & \multicolumn{3}{c|}{20170929} & \multicolumn{3}{c}{20170930}  \\
\cline{2-10}
MAPE(\%) & 5.113    & 20.554  & 22.022  & 4.529    & 28.769  & 20.387  & 5.248   & 39.266   & 58.505   \\
RMSE     & 219.126  & 658.361 & 588.094 & 211.379  & 840.249 & 680.796 & 239.917 & 1578.820 & 1114.432 \\
R-square & 0.997    & 0.977   & 0.982   & 0.998    & 0.966   & 0.978   & 0.996   & 0. 816   & 0.909   \\
\hline
\end{tabular}
\label{tab_4}
\end{table}

The results from network-wide perspective are illustrated in Figure~\ref{fig_8} and Table~\ref{tab_4} for three models on network-wide results. For any toll station, the average of metric MAPE over the days in test data is calculated, and is drawn as Figure~\ref{fig_8} with the same notions of Figure~\ref{fig_6}. Our method performs the best again due to visible positive-skew distribution. Besides the similar conclusions from Figure~\ref{fig_6}, we found another two facts. First, in the distribution of GCN and GTM, there are not many counts in the histogram’s first bucket (i.e., with the lowest error). The graphs of both models are built by physical highway network whose weights based on static distances. However, such geographic proximity doesn’t directly affect highway traffic volumes~\cite{article_3}: a vehicle always runs long distance in highway network, may not exit at a toll station near from the entry one. While in phdST, such spatial influences are modelled by upstream dependent stations in Definition 3. Second, GTM performs a little better than GCN in general, because more counts lies in the first three buckets. It comes from GTM considers more in temporal dimension. Furthermore, the same six specific days as Experiment 1 are used for detailed statistics as Table~\ref{tab_4}. From metrics R-square and RMSE, GTM is proved a little better than CCN on any of the six days; while from metric MAPE these graph based models have their own merits. Our phdST performs the best of all, which comes from phased deep learning.

\begin{figure}[h]
  \centering
  \includegraphics[width=\linewidth]{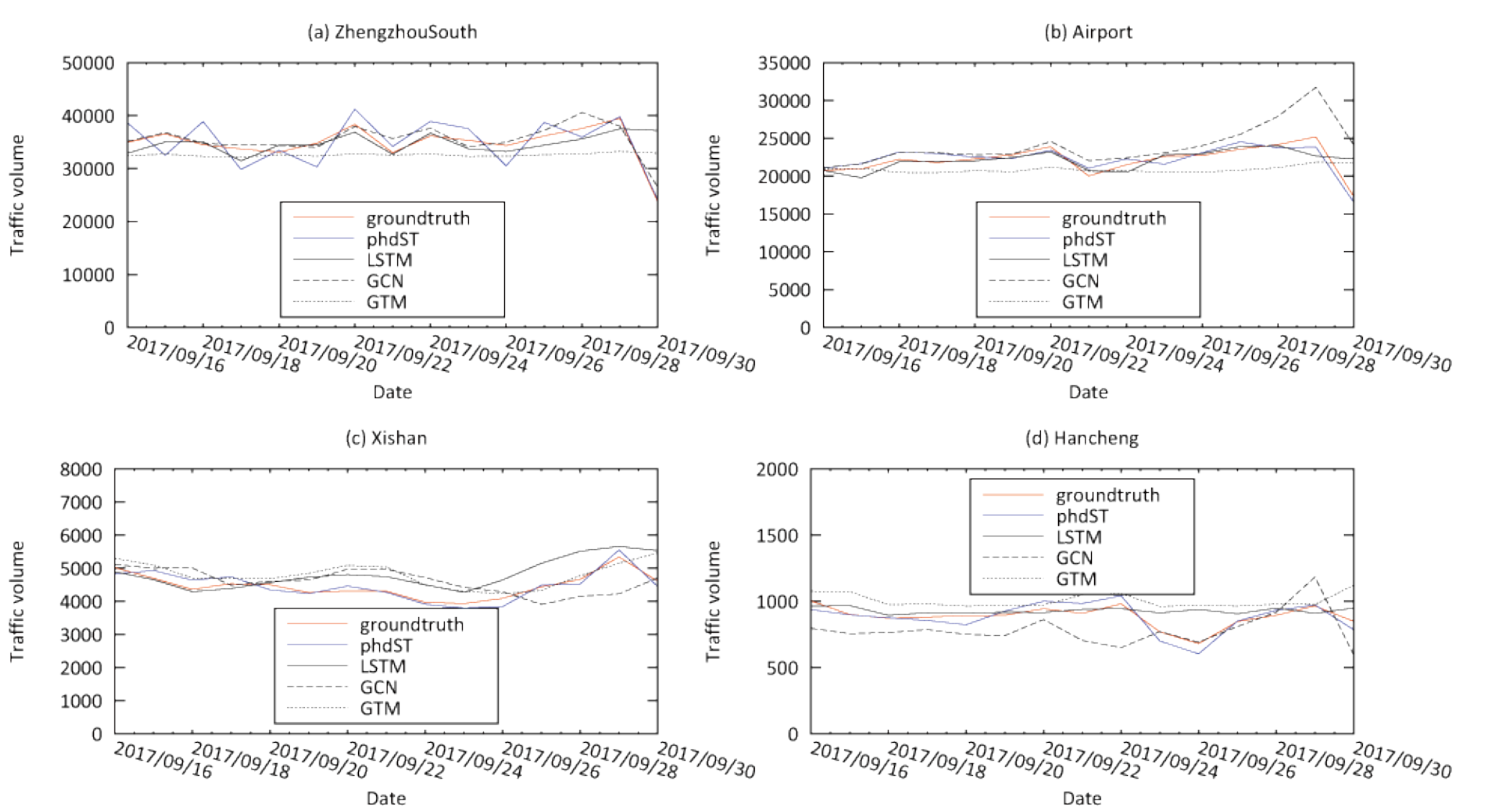}
  \caption{Traffic volume in station perspective.}
  \label{fig_9}
\end{figure}

The results from station perspective are represented in Figure~\ref{fig_9} and Table~\ref{tab_5}, in which the same four stations on the same 15 days as experiments above are used. As Figure~\ref{fig_9}, at each of those four toll stations, the predicted traffic volumes of four models fit the ground truth values, and phdST works better to depict fluctuations of daily traffic volumes. LSTM and two graph based models seem to perform not well at the toll stations with small traffic, where seldom fluctuation is reflected. Further details can be found from the statistics in Table~\ref{tab_5}. LSTM appears wider variation especially at the toll stations with small traffic. GTM achieves obviously good results only at the moderate stations (e.g., xishan), and ARIMA cannot work well at toll stations with small traffic volumes. With the help of phased deep learning our method performs the best at any toll stations again with steady performance.

\begin{table}[]
\caption{Prediction performance in station perspective.}
\begin{tabular}{c|cccc|cccc}
\hline
         & phdST    & LSTM     & GCN      & GTM      & phdST   & LSTM     & GCN      & GTM      \\
\hline
         & \multicolumn{4}{c|}{ZhengzhouSouth}        & \multicolumn{4}{c}{Airport}              \\
\cline{2-9}
MAPE(\%) & 7.297    & 7.345    & 3.686    & 10.264   & 3.364   & 4.337    & 8.804    & 8.793    \\
RMSE     & 2948.135 & 3787.689 & 1510.562 & 4081.028 & 813.132 & 1532.830 & 2802.311 & 2269.452 \\
R-square & 0.952    & 0.921    & 0.987    & 0.909    & 0.987   & 0.954    & 0.847    & 0.899    \\
\hline
         & \multicolumn{4}{c|}{Xishan}                & \multicolumn{4}{c}{Hancheng}             \\
\cline{2-9}
MAPE(\%) & 3.369    & 9.214    & 9.863    & 8.699    & 4.997   & 7.990    & 14.605   & 14.443   \\
RMSE     & 168.6719 & 488.372  & 526.796  & 453.344  & 51.191  & 88.985   & 163.239  & 143.896  \\
R-square & 0.985    & 0.878    & 0.858    & 0.895    & 0.972   & 0.915    & 0.712    & 0.776   \\
\hline
\end{tabular}
\label{tab_5}
\end{table}

\section{Conclusion}
In this paper, a phased deep learning method is proposed in a business interpretable way for daily traffic volume prediction in highway. A hybrid model combining FCN and LSTM in spatio-temporal learning phase fully considers time, space, meteorology, and calendar on heterogeneous spatio-temporal data to improve predictive accuracy. Meanwhile, daily traffic volumes especially at vital few toll stations are calibrated effectively by feature pre-processing phase and decision phase. Evaluated on the real-world data in a practical project, performance advantages and convincing benefits are proved by extensive experiments.

Appealing graph characteristics of highway are expected to be employed in our future work. Accordingly, the trade-off among expressive spatial characteristics, relative high complexity, and long training latency, has to be considered next in graph convolution neural network.

\section{Acknowledgments}

This work was supported by Beijing Municipal Natural Science Foundation (No. 4202021), Top Young Innovative Talents of North China University of Technology (No. XN018022), and “Yuyou” Talents of North China University of Technology (No. XN115013).

\bibliographystyle{ACM-Reference-Format}
\bibliography{phdST_REF}


\end{document}